# Feature Extraction, Classification and Prediction for Hand Hygiene Gestures with KNN Algorithm

Rashmi Bakshi

*School of Electrical and Electronic Engineering, TU Dublin-City campus*

**Abstract**

There are six, well-structured hand gestures for washing hands as provided by World Health Organisation guidelines. In this paper, hand features such as contours of the hands, the centroid of the hands and extreme hand points along the largest contour are extracted for specific hand-washing gestures with the use of a computer vision library, OpenCV. For this project, a robust dataset of hand hygiene video recordings is built with the help of 30 research participants. In this work, a subset of the dataset used as a pilot study to demonstrate the effectiveness of the KNN algorithm. Extracted hand features saved in a CSV file are passed to a KNN model with a cross-fold validation technique for the classification and prediction of the unlabelled data. A mean accuracy score of >95% is achieved and proves that KNN algorithm with an appropriate input value of K=3 is efficient for hand hygiene gestures classification.

## 1   Introduction

The current ongoing coronavirus pandemic has highlighted the importance of hand hygiene practices in our daily lives, with governments and healthcare authorities around the world promoting good hand hygiene practices. In Europe, cases of hospital-acquired infections (HAIs) are more than 1 million annually [1]. Hand hygiene compliance may reduce the risk of cross-transmission of microbes such as MRSA, bacteria commonly known for causing the spread of HAIs thereby increasing the number of infections as well as health-care expenditures [2]. There are well-documented and structured guidelines for washing hands as per World Health Organization (WHO) [3]. In advance of developing, a real time hand-hygiene gesture detection system, a preliminary analysis for feature extraction associated with each hand gesture is required. In this paper, OpenCV library is used to extract features for hand hygiene gestures such as- "rub hands palm to palm", "hands with fingers interlocked" and "hands with fingers interlaced". The features further passed to a KNN model for hand hygiene gesture classification and prediction.

## 2 Hand Hygiene Gesture Dataset

Thirty volunteers participated in the process of data collection to perform the various hand washing gestures as per WHO guidelines. An aluminium rig is placed over the laboratory's sink, incorporated with a digital camera for recording of the hand gestures. The camera device captures hand movements in isolation thereby protecting the user's privacy. The participants completed an information sheet along with the consent form for their inclusion in the research. The participants saw a visual demonstration of WHO hand hygiene stages before recording their video data. They originated from diverse ethnic backgrounds thereby providing skin colour variation in the dataset. A robust dataset of 30 videos-six hand hygiene gestures is developed. Skin extraction algorithm (YCbCr color space) to extract hands for users with different skin colour tones is implemented. The video length for each hand washing activity was 25-30 seconds. Every hand-washing step was followed by a pause where in the user was instructed to move their hands away from the camera. Video format for this data set is MP4 file with a size of range 40-60 MB and a frame rate of 29.84 frames/s. All of the six hand washing movements were recorded in one video for each participant. In this paper, a subset of the data set is utilised with an aim for the use of a complete dataset in future studies [4].

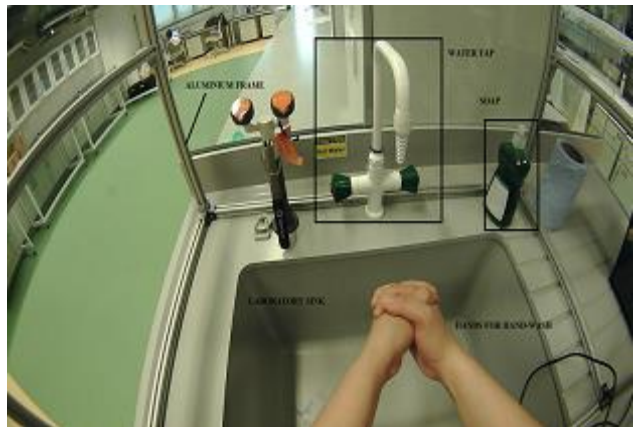

**Figure 2: An Aluminium frame placed onto the laboratory sink for data collection**

## 3 Edge Detection in Computer Vision

A contour is a closed curve of points or line segments that represents the boundaries of an object in an image, having the same colour and intensity [5]. Finding contours determines the shape of an object (arc length; number of vertices), the number of objects in an image (number of contours) and measure the size of the objects [5].

Contour features: Contour features are the attributes of a contour derived from the contour, used for feature extraction and classification of the images. The common contour features are:

1. Image Moments: Image moments are statistical properties of a section of an image. Image moments are used to extract useful information from the contour such as the centroid, area, etc. The moments are known as features for shape recognition.

2. Centroid: The centroid is defined as a coordinate (cx, cy) and is derived from the image moments.

3. Contour area: The contour area is the image area outlined by the contour.

4. Contour perimeter: is also called the 'arc length'; it is the length of the contour in pixels.

5. Convex hull: The convex hull of a shape or a group of points is a tight fitting convex boundary around the points or the shape of the object tracked. They are the minimum enclosing polygon of all points of the input shape. With a given set of X points in the Euclidean space, the convex hull is the smallest possible convex set containing these X points [6, 7].

Contour tracking is widely used in the field of computer vision. The idea behind contour tracking is to traverse the border of a region completely and detect the edge points. Xie et al. [8] has used the contour detection method to determine the number of copper cores in the wire. Poda et al. [9] has found the perimeter of a contour to detect a specific shape such as a pentagon, and the initial character 'P' is saved and transmitted to an Arduino for the movement of a mechanical arm. Bochkarev et al. [10] compares object characteristics such as area, perimeter and compactness of the contour of regular shapes to that of irregular shapes using Computer vision-open source library-OpenCV. OpenCV offers cv2.findContours function, which can retrieve contours from the binary image and return the number of detected contours. The hand contours and centroid are utilised as 'features' in determining the shape based hand gestures by calculating the Euclidean distance between centroid and hand contours [11, 12].

In this work, RGB image is converted to YCbCr color space, creating a mask with widely accepted values for skin detection. Then, contours for the skin pixels are retrieved. The centroid of the hands with the help of image moments are detected. A Euclidean distance between the centre point and contours is calculated to determine the extreme hand points. Argmin and argmax functions to slice a numpy array are used to extract the extreme points for each frame. Centroid and extreme hand points are saved in a csv file as a tuple for each hand hygiene video recording for further processing and classification.

## 4  K-Nearest-Neighbours Algorithm

The K-Nearest-Neighbours (KNN) is a nonparametric classification algorithm, i.e. it does not make any presumptions on the given dataset. It is known for the simplicity and effectiveness [13]. It is a supervised learning algorithm where a training dataset is provided and the data points are categorised into pre-defined classes. The training of the model is carried out to predict the class of an unlabelled data. KNN has a high cost of classifying new instances as nearly all computation takes place at classification time rather than when the training examples are first encountered and therefore also known as a lazy learning algorithm. It is used to classify the data based on closest or neighbouring training examples in a given region. For continuous data, Euclidian distance between the data points is calculated to determine the closest neighbours. The input value of 'K' is used to determine the number of neighbours and to build the boundaries for each class. It is usually combined with cross validation method, which is a resampling procedure and splits the data in K number of groups [13, 14]. In this work, K-fold cross validation is implemented in python as an adaptation from [15].

*System requirements*

The hardware used for all the experiments were Intel(R) Core(TM) i5-5200U CPU @ 2.20GHz 2.19GHz with 4 GB RAM with 64-bit Windows Operating System. The python environment with OpenCV and scikit-learn library packages are utilised for conducting the experiments.

## 5  Methodology

The methods can be sub divided into four sections. Skin detection is the first step where hands are extracted for each frame in a video recording. Contour-Centroid detection is carried out for each hand hygiene pose accompanied by extreme hand locations. These features are saved as a tuple in a csv file and passed on to a KNN classification model for the training and prediction of the unlabelled row of data.

5.1  *Skin Detection Algorithm:*

Convert RGB image frames in a video to YCbCr color space. Apply the skin mask of given values to extract hands in each frame.

    min_YCrCb = np.array([0,133,77],dtype="uint8"); max_YCrCb = np.array([255,179,127],dtype="uint8")

5.2  *Centroid of the hand:*

Centroid of a two-dimensional shape is the average position of all the point in the shape. Retrieve image moments for the largest contour and compute X, Y coordinates for the centroid.

    $c_X = int(M["m_{10}"] / M["m_{00}"])$ ; $c_Y = int(M["m_{01}"] / M["m_{00}"])$ ; centeroid=$(c_X, c_Y)$

Figure 5.2 is a sample image outlined with hand contours and a centroid encircled.

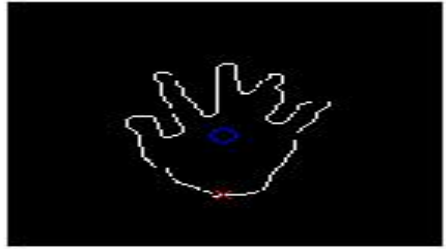

**Figure 5.2 a hand skeleton outlined with hand contours and a circular centroid.**

## 5.3 Extreme hand points:

The extreme hand points (top, bottom, left and right) along with the largest contour are extracted for each frame in a hand hygiene video recording. The extracted features are saved in hand_features.csv file.

The distance from the centroid and contour points is determined by:

$$d = \sqrt{(c_x - x_i)^2 + (c_y - y_i)^2} \qquad (1)$$

In equation (1) $d$ is the distance, $c_x$ is x-axis coordinate of centroid, $c_y$ is y-axis coordinate of the centroid. $x_i$ is x-axis of detected hand contour and $y_i$ is y-axis coordinate of detected hand contour. In python, argmin and argmax functions are used to extract the extreme points. Figure 5.4 a) elaborates all the steps for feature extraction.

## 5.4 KNN model:

Features saved in a csv file are passed to a KNN model that determines the nearest neighbours with the value of K=3. The set of observations is randomly divided into K groups or folds of approximately an equal size. The first fold serves as a validation set and the method is fit on the remaining K-1 folds. The mean evaluation score is given for each fold in Table 6.2. Figure 5.4 b) presents the sequential workflow for training of the dataset; classification and prediction of the unlabelled row of data.

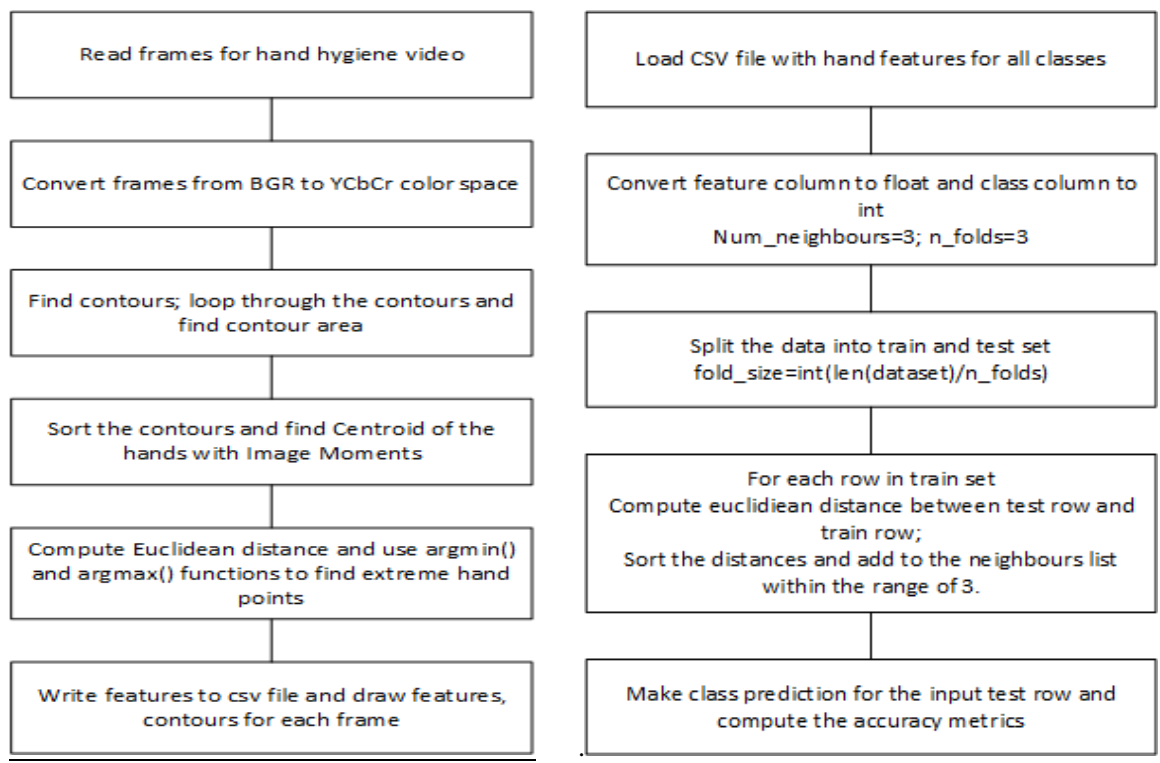

**Figure 5.4 a): Hand Features Extraction Steps**   **Figure 5.4 b): Workflow to train KNN Model with hand features**

# 6 Results

This section provides the results for feature extraction and the classification of the unlabelled data. A sample of the extracted features for a given hand hygiene video clip is presented. Centroid, extreme hand points- left, right, top and bottom- x, y coordinates are saved as a tuple in a csv file. The class name is added manually in the csv file for each row of features.

```
Centroid: (112,175); Extreme Left:( 74,192); Extreme Right: (153,149);
Extreme Top: (122,104); Extreme Bottom (74,239) Class: Palm to Palm
```
**Figure 6.1: A sample of features saved in Hand_features.csv**

Table 6.1 lists three hand hygiene classes corresponding to three hand hygiene videos that are utilised from a robust hand hygiene dataset, discussed in section 2. Number of extracted feature rows for each video clip is given.

| Class Label | Number of rows for Extracted Features saved in a csv file |
|---|---|
| Palm to Palm | 173 |
| Fingers Interlaced | 42 |
| Fingers Interlocked | 149 |

**Table 6.1: Number of features- rows extracted for each hand hygiene class**

Figure 6.2 is the result of image processing where the RGB pixels are converted to YCBCR color space for skin detection with minimum (0, 133, 77) and maximum (255, 179, 127) range used as a skin mask. The largest contour max (contour area) is displayed along with the contour features such as centroid and extreme hand points along the contour.

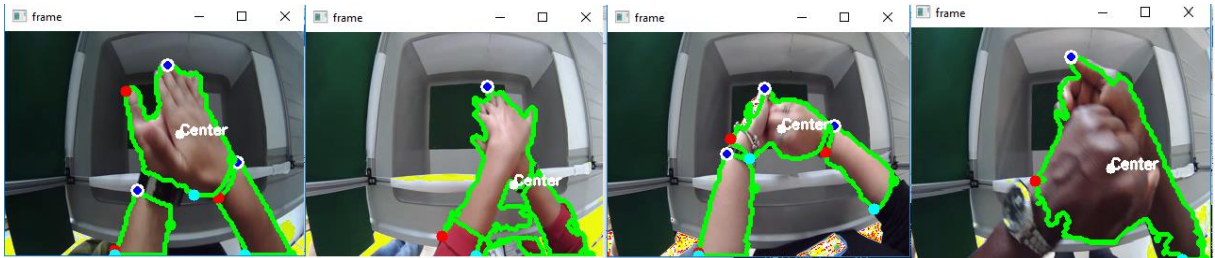

**Figure 6.2: Sample frames for three different hand hygiene gestures with extracted features (Centroid, extreme hand points)**

Figure 6.3 a) represents the sample of hand features- Centroid and extreme finger position TOP for 3 different classes- Palm to Palm, Fingers Interlocked and Fingers Interlaced. The normalized features within the range of {0-1} are represented in Figure 6.3 b). In an ideal case with a static setup, data points- features belonging to the same class- (same colour code) must be equal to each other for an accurate working of the KNN algorithm. However, it is unrealistic for all data points to be exact same and therefore K=3 is selected to determine the three closest neighbours.

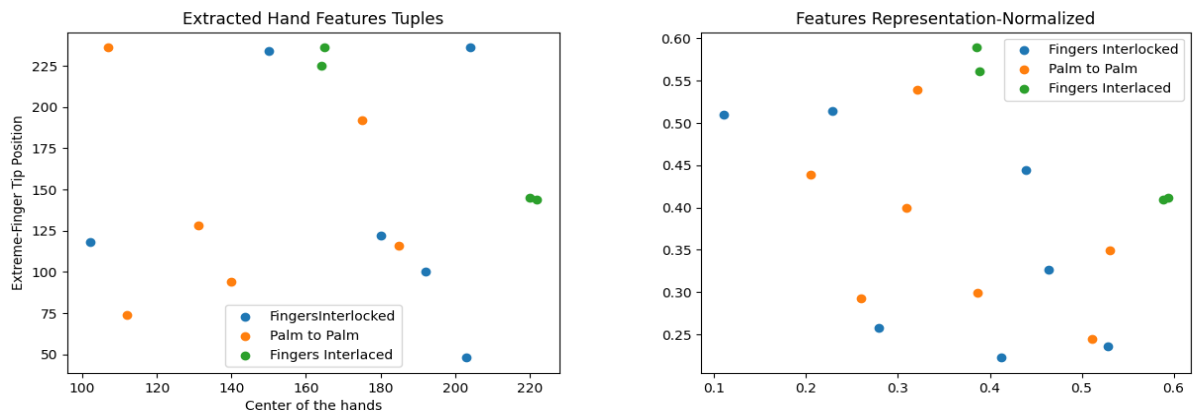

**Fig 6.3 a) Hand features for distinct hand hygiene classes   Fig 6.3 b) Centroid and extreme hand points- normalised**

K Nearest Neighbours (KNN) algorithm classifies the data based on the similarity found by the Euclidian distance among the data points. Similar and closest data points are grouped together in one class. The success rate of the algorithm is dependent on the input value of K. An appropriate value of K with cross-fold validation can increase the accuracy of the classifier. The accuracy score for each fold along with a mean accuracy metric is given in a Table 6.2.

A new data frame is selected from the hand hygiene video dataset that the model has not encountered before. Features for the subsequent frames: centroid and the extreme points are passed as an input for the class prediction. Table 6.3 lists the unlabelled data frames with class predictions and it is observed that KNN correctly classifies 6 out of 7 rows of features.

| Accuracy Score | Fold 1 | Fold 2 | Fold 3 |
|---|---|---|---|
| (Correctly Predicted/Actual)*100 | 95.87 | 96.69 | 96.69 |
| **Mean Accuracy** | 96.41% | | |

**Table 6.2 Accuracy score for each fold passed to KNN algorithm with input K=3**

| Unlabelled Features | Predicted Class K=3 | Actual Class | Correct/Incorrect Prediction |
|---|---|---|---|
| [112,175,74,192,153,149,122,104,74,239] | Palm to Palm | Palm to Palm | ✓ |
| [180,102,122,118,231,108,162,54,148,138] | Fingers Interlocked | Fingers Interlocked | ✓ |
| [192,204,100,236,299,239,158,144,100,239] | Fingers Interlaced | Fingers Interlocked | x |
| [159,19,96,32,248,13,112,0,96,71] | Fingers Interlocked | Fingers Interlocked | ✓ |
| [140,131,94,128,189,97,174,42,140,221] | Palm to Palm | Palm to Palm | ✓ |
| [220,164,145,225,299,239,194,60,180,239] | Fingers Interlaced | Fingers Interlaced | ✓ |
| [185,107,116,236,177,205,141,168,116,239] | Palm to Palm | Palm to Palm | ✓ |

**Table 6.3 Class Prediction for the unlabelled data frame**

# 7     Conclusion

This paper extracts hand features- centroid, contours and extreme hand points with a computer vision library- OpenCV. It proves the effectiveness of KNN algorithm for hand gesture classification by finding the closest data points in a training dataset. This work attempts to answer the following questions with respect to gesture classification research.

*Why contours- (centroid and extreme hand points) information is important for gesture classification.*

Extreme hand points correspond to the corners in an image. In computer vision, corners are a strong feature as they are uniquely identifiable due to large variations in all directions. Each hand hygiene pose is unique. Their Centroid feature is distinct, local, accurate, and therefore suitable for hand gesture classification.

Silvia et al. [11] has implemented a hand gesture recognition for sign language gestures by calculating the Euclidean distance between hand contours.

*List the strengths and weakness of the hand –hygiene dataset constructed for this research.*

The hand hygiene dataset includes hand washing video recordings from 30 participants with diverse ethnicity. The strength lies in having the skin color variation and therefore a robust dataset wherein the personalised interpretation of the hand hygiene guidelines for washing hands were carried out. This dataset can be further explored as a knowledge base to carry out research in multi-disciplinary fields such as human factors, human computer interaction and computer vision. The dataset recorded in a laboratory setting, can be viewed as a limitation as all of the video clips are captured through a stationary camera at a fixed angle. Future data collection will focus upon camera rotation and different lighting conditions to compose robust images and extract invariant features.

*Why was KNN selected as a classification learning algorithm with input value of K= 3*

K Nearest Neighbours (KNN) algorithm classifies the data based on the similarity found by the Euclidian distance among the data points. Each hand hygiene pose is unique with distinct centroid and hand extremity, maxima – finger locations. KNN is suitable to classify hand hygiene gestures as determines the closest neighbours to the given point of interest. Repetitive experiments were conducted with an input value of K=2, K=3 & K=5 for the prediction of the unlabelled data. K=2 neighbours caused over fitting; performed well on the training set but gave incorrect predictions for the unseen data. K=5 caused under fitting with smoother boundaries and generalised model and therefore an input value of 3 was finally selected for prediction.

*List the future pathway for this research project.*

Answer : In this work, KNN algorithm has proved to be an efficient supervised learning technique in hand hygiene gesture recognition with a mean accuracy > 95% for the three distinct hand hygiene stages- "rub hands palm to palm", "hands with fingers interlocked" and "hands with fingers interlaced". This research will expand in order to incorporate all of the six hand hygiene stages along with the use of a complete hand gesture dataset. Inclusion of a revised dataset with a different camera setup and room lighting to allow image variation.

*Supplementary Materials*: The hand hygiene video recordings along with the python code; hand-features-csv file for the following are available online:
https://tudublin-my.sharepoint.com/:f:/r/personal/d16126930_mytudublin_ie/Documents/Hand%20Hygiene%20Research/KNNClassification?csf=1&web=1&e=EFHrc9
The hand hygiene dataset is available:
https://tudublin-my.sharepoint.com/:f:/r/personal/d16126930_mytudublin_ie/Documents/Hand%20Hygiene%20Research/HandWashData?csf=1&web=1&e=mMwzfp.